%% file: main.tex
\documentclass{article}

\usepackage[numbers,square]{natbib}
\usepackage[nonatbib,final]{corl_2022}

\usepackage[utf8]{inputenc} % allow utf-8 input
\usepackage[T1]{fontenc}    % use 8-bit T1 fonts
\usepackage{hyperref}       % hyperlinks
\usepackage{url}            % simple URL typesetting
\usepackage{booktabs}       % professional-quality tables
\usepackage{amsfonts}       % blackboard math symbols
\usepackage{nicefrac}       % compact symbols for 1/2, etc.
\usepackage{microtype,wrapfig}      % microtypography

\usepackage{microtype,bbm}

\usepackage{algorithm}
\usepackage[noend]{algpseudocode}
\newcommand\NoThen{\renewcommand\algorithmicthen{}}
\usepackage{graphicx}
\usepackage{booktabs,makecell,tabularx} 
\usepackage{hyperref}
\usepackage{pdfpages}
\usepackage{subcaption}
\usepackage{mwe}

\usepackage{float}
\usepackage{amsmath,amssymb,amsthm}
\usepackage{todonotes}

\DeclareMathOperator*{\argmaxA}{arg\,max}

\newtheorem{definition}{Definition}

\newtheorem{proposition}{Proposition}

\usepackage{pdfpages} % include pdfs
\usepackage{pgffor} % for loops
\makeatletter
\AtBeginDocument{\let\LS@rot\@undefined}
\makeatother
\def\supplementfilename{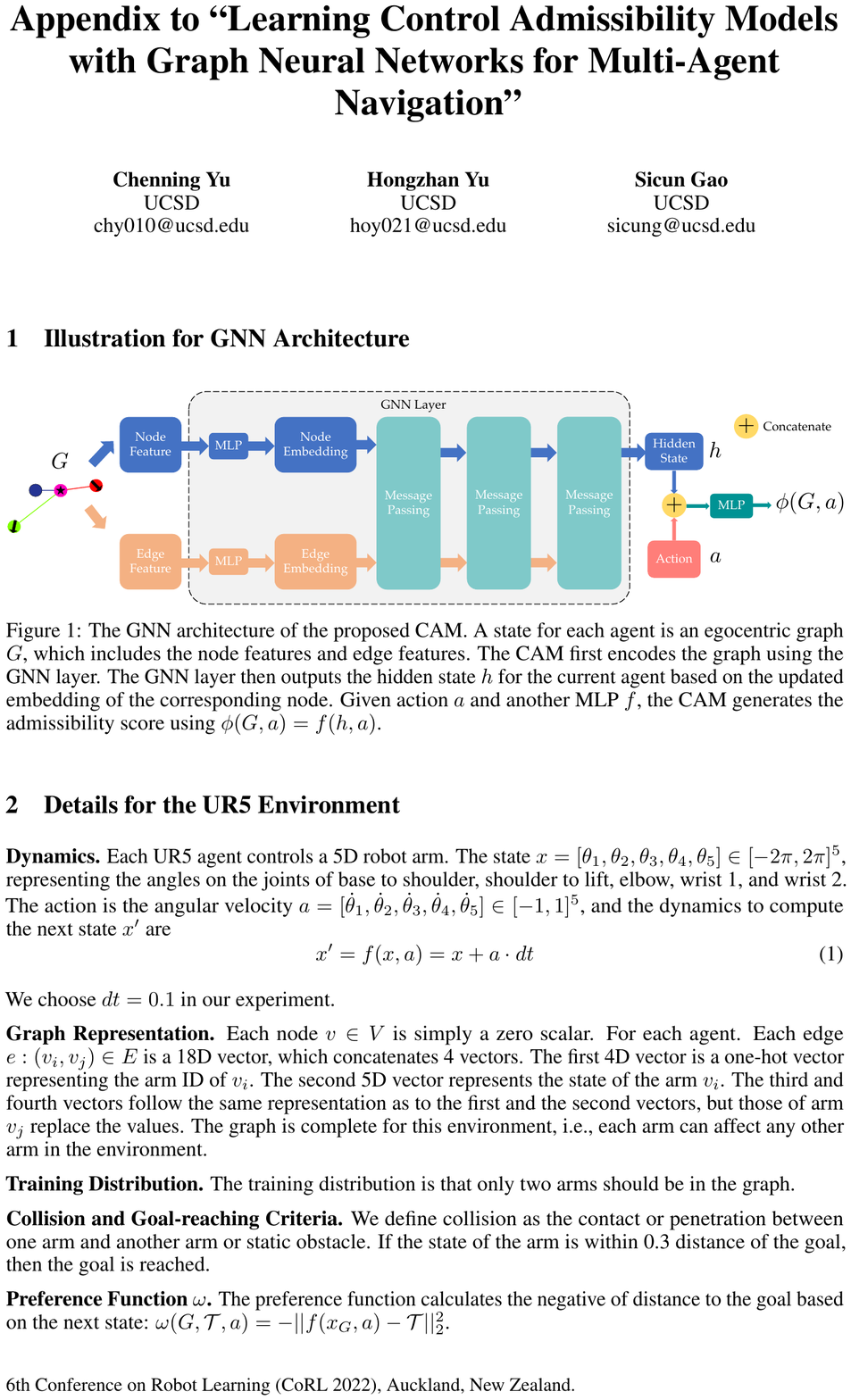}
\pdfximage{\supplementfilename}
\def\numbersupplementpages{\the\pdflastximagepages}
\newif\ifarXiv
\arXivtrue 
\begin{document}

\title{Learning Control Admissibility Models with Graph Neural Networks for Multi-Agent Navigation}

\author{Chenning Yu\\UCSD\\chy010@ucsd.edu\And Hongzhan Yu\\UCSD\\hoy021@ucsd.edu\And Sicun Gao\\UCSD\\sicung@ucsd.edu}

\maketitle

\begin{abstract}
Deep reinforcement learning in continuous domains focuses on learning control policies that map states to distributions over actions that ideally concentrate on the optimal choices in each step. In multi-agent navigation problems, the optimal actions depend heavily on the agents' density. Their interaction patterns grow exponentially with respect to such density, making it hard for learning-based methods to generalize. We propose to switch the learning objectives from predicting the optimal actions to predicting sets of admissible actions, which we call control admissibility models (CAMs), such that they can be easily composed and used for online inference for an arbitrary number of agents. We design CAMs using graph neural networks and develop training methods that optimize the CAMs in the standard model-free setting, with the additional benefit of eliminating the need for reward engineering typically required to balance collision avoidance and goal-reaching requirements. We evaluate the proposed approach in multi-agent navigation environments. We show that the CAM models can be trained in environments with only a few agents and be easily composed for deployment in dense environments with hundreds of agents, achieving better performance than state-of-the-art methods. 
\end{abstract}

\section{Introduction}

Multi-agent navigation is a longstanding problem with a wide range of practical applications such as in manufacturing~\cite{manufact1,manufact2}, transportation~\cite{transport1}, and surveillance~\cite{survei1}. The goal is to control many agents to all achieve their goals while avoiding collisions and deadlocks. Such problems are multi-objective in nature, and the control methods depend heavily on the density of the agents, with the computational complexity growing exponentially in such density~\cite{masurvey1,masurvey2,panait2005scalability,brambilla2013scalability}.

Recently, reinforcement learning approaches have shown promising results on multi-agent navigation problems~\cite{macommunicate,gpg,PIC,maddpg,QMIX}, but there are still two well-known sources of the difficulty. First, balancing safety and goal-reaching requirements often lead to ad hoc reward engineering that is highly dependent on the environments and the density of the agents~\cite{rewardhacking1,rewardhacking2}. Second, it is shown that such neural control policies leaned on RL approaches are hard to generalize when the agent density in the environment changes, which leads to distribution drifts. 
The core difficulty for generalizability is that the RL approaches are designed to capture the optimal actions specific to the training environments and rewards. Nevertheless, such optimal actions may change rapidly when the density and interaction patterns change during deployment. The unnecessary optimization of control policies in the training environments makes it fundamentally hard to transfer the learned results to new environments and achieve compositionality in the deployment to varying numbers of agents. 

\begin{figure*}[!t]
\centering
\includegraphics[width=\textwidth]{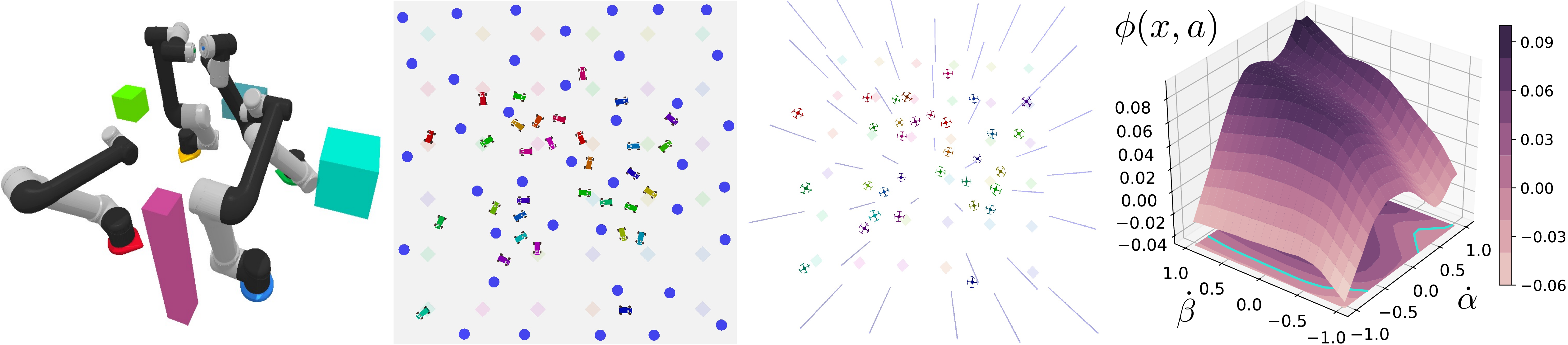}
\text{\hspace{20pt}(a) ~UR5\hspace{55pt}(b) ~Car\hspace{58pt}(c) ~Drone\hspace{40pt}(d) ~Landscape of CAM}
\caption{\textbf{(a-c)} Illustrations of the multi-agent environments. We show videos in the supplemental materials. \textbf{(d)} An example landscape of the proposed CAM $\phi$ in the Drone environment. The $\dot\alpha$-axis and $\dot\beta$-axis, are two dimensions in the action space, corresponding to the rotation rates in pitch and roll. We project the landscape into a 2D plane at the bottom. The blue line on the plane denotes the zero level set, which divides the action space into the admissible and inadmissible sets.}
\label{environments}
\end{figure*}

We propose a new learning-based approach to multi-agent navigation that shifts the focus from learning the optimal control policy to a set-theoretic representation of {\em admissible} control policies to achieve compositional inference. 
\textcolor{black}{By decoupling the multi-agent navigation tasks, we avoid the ad-hoc reward engineering using a simple goal-reaching preference function and a learnable set-theoretic component for collision avoidance.}
We use graph neural networks (GNNs) to represent the set of admissible control actions at each state as Control Admissibility Models (CAM), which are trained with a small number of agents during training with sparse rewards. In online inference, we compose the CAMs and apply goal-reaching \textcolor{black}{preference functions} to infer the specific action to take for each agent. We show that CAMs can be learned purely from the data collected online in the standard RL setting, with no expert or human guidance. 
We propose relabelling through backpropagation (Section \ref{method:relabel}) for the CAMs to learn rich information from the transitions even given sparse rewards. As shown in Figure \ref{environments}(d), the learned models of the admissible actions are typically complex and suitable for GNN representations. Moreover, in online inference, we decompose the state graph of the agent into much smaller subgraphs and aggregate the CAMs from them. Such compositionality allows us to train CAMs with only a small number of agents, directly deploy them in much denser environments, and achieve generalizable learning and inference. 

We evaluate the proposed methods in various challenging multi-agent environments, including robot arms, cars, and quadcopters. Experiments show that the CAM generalizes very well. While the training environments only have at most 3 agents, the learned CAM models can be directly deployed to hundreds of agents in comparable environments and achieve a very high success rate and low collision rate. We analyze the benefits of the proposed methods compared to state-of-the-art approaches. Furthermore, we show a zero-shot transfer to a multi-agent chasing game, demonstrating that the trained CAM generalizes to other tasks that are not limited to navigation. 

\noindent{\em \textbf{Related Work.}} \input{related}

\begin{figure*}[!t]
\centering
\includegraphics[width=0.98\textwidth]{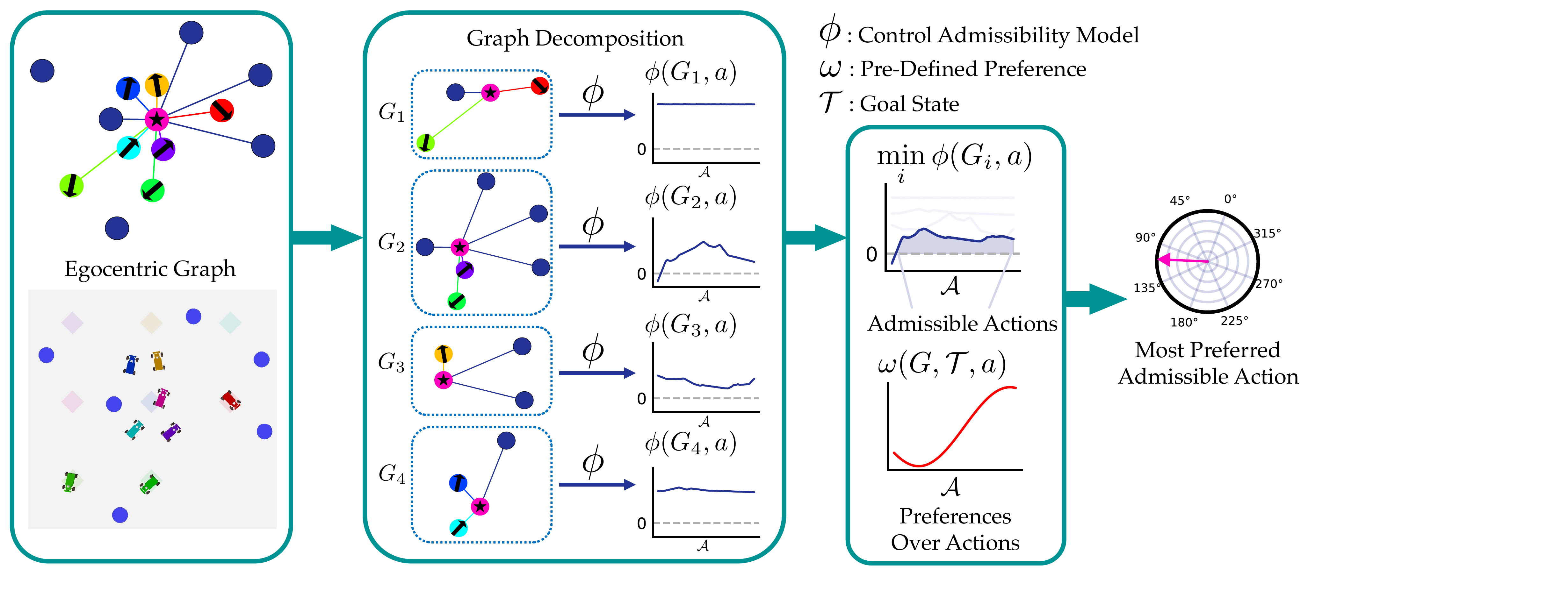}
\caption{An overview of the proposed CAM approach. At each time step, the state of each agent is an egocentric graph. The graph is further decomposed at inference time into a set of subgraphs to follow the training data distribution. The CAM takes these subgraphs and outputs the admissibility scores along with a set of sampled candidate actions. We filter out the inadmissible actions by checking whether the minimum of the scores is below 0. Given a predefined preference over actions, the model outputs the most preferred action in the admissible set for each agent.}
\label{framework}
\end{figure*}

\section{Control Admissibility Models}

We present Control Admissibility Models (CAMs), a general representation of the feasible control set. CAM is task-compositional, robust to reward design, and can be trained with an online pipeline similar to model-free RL. With graph decomposition, CAM can plan a high-quality solution for a large number of robots, which could be significantly out of the distribution of the training tasks.

\subsection{Control Admissibility Models (CAMs)}

We define CAMs as a general representation of the feasible actions. In general, CAMs are function approximators that behave as the characteristic functions of such sets. By evaluating the CAM values we can efficiently determine whether an action is feasible. Formally, 
\begin{definition}[Control Admissibility Models and Admissible Set] Given CAM $\phi$, state $x$, action space $\mathcal{A}$, the admissible set is defined as $\Delta(\phi,x): \{a\mid a\in\mathcal{A},\phi(x,a)\geq 0\}$.

\end{definition}

\noindent{\bf Compositionality of CAMs.} An important feature of CAMs is that they can be composed together. Given two CAMs $\phi_1$ and $\phi_2$, we want to find the admissible actions that satisfy both these two CAMs: \[\{a \mid \phi_1(x, a)\geq 0 \} \cap \{a \mid \phi_2(x, a)\geq 0 \}\]

\begin{proposition}[Composition]\label{composition}
If $\phi_1$ and $\phi_2$ are CAMs, then $\Delta(\phi_3, x)$ defines $\Delta(\phi_1,x)\cap \Delta(\phi_2,x)$, given $\phi_3=\min\{\phi_1, \phi_2\}$. 
\end{proposition}

\subsection{Graph Neural Networks for CAMs}\label{alg::gnn_for_cam}

Though CAM can be applied to various single-agent problems (see Section \ref{singleagentenv} as an example), we focus on multi-agent problems in this paper. At each timestep, the state for each agent is an egocentric graph $G=\langle V, E \rangle$, where $V$ are the vertices and $E$ are the edges.
Vertices $V$ is composed of agent vertices $V_a$ and static obstacle vertices $V_o$. Edges $E$ are connecting all the neighbor vertices to the current agent, i.e. $E=\{(v_a^{j},v_a^{i}) | a_j \in \mathcal{E}(a_i)\cup\{(v_o^{j},v_a^{i}) | o_j \in \mathcal{E}(a_i)\}$, where $\mathcal{E}$ denotes the neighbor vertices. The one-hot vector is used as the features of $V$, denoting the types of vertices. For the features of edges $E$, we use relative positions and the states of the two connected vertices, along with a one-hot vector to indicate the edge types. Zero paddings are adopted if necessary~\cite{alexnet}. We provide more details on the representation of graphs in the Appendix.

The CAM $\phi$ has two components: The first component is a GNN layer, which transforms the graph $G$ to a hidden state vector $h$ for each agent. The second component is a fully-connected layer $f$, which takes the hidden state $h$ and an action $a$, and predicts the admissibility score $\phi(G, a)=f(h, a)$. Such a design enables fast computation of the admissibility scores for all the possible state-action pairs. We can merge the computation of the hidden state for all state-action pairs which share the same input graph into one forward passing in the GNN layer.

We describe the GNN layer with the following details. The vertices and the edges are first embedded into latent space as $m^{(0)}=g_m(v), n^{(0)}=g_n(e)$, given fully-connected layers $g_m, g_n$. Taking $m, n$, the GNN aggregates the local information for each vertex from the neighbors through $K$ GNN layers. For the $k$-th layer, it performs the message passing with 2 fully-connected layers $f_m^{(k)}$ and $f_n^{(k)}$:
\begin{equation}
\begin{split}
m_i^{(k+1)}&=m_i^{(k)} +\max\{f_m^{(k)}(n_l^{(k)})\mid e_l:(v_j,v_i)\in E\}, \forall v_i \in V\\
n_l^{(k+1)}&= n_l^{(k)} + f_n^{(k)}(m_i^{(k+1)}, n_l^{(k)}), \forall e_l:(v_j,v_i) \in E
\label{coreupdate}
\end{split}
\end{equation}
We assign the hidden state $h$ with the final node embedding $m_i^{(K)}$, where $i$ corresponds to the current agent. We use $\max$ as the aggregation operator to gather the local geometric information due to its empirical robustness to achieve the order invariance~\cite{macbf,pointnet}. We use the residual connection to update the vertex embedding and edge embedding due to its simplicity and robust performance for deep layers~\cite{resnet,resgnn}. Moreover, each agent's hidden state $h$ is only updated based on its egocentric graph, which is entirely invariant to the other agents' hidden states. This architecture decentralizes the decision-making of the whole swarm.

\subsection{Training CAMs in Reinforcement Learning}

We train CAMs in the same online setting as model-free reinforcement learning procedures. At each time step, the agent perceives the observation from the environment and takes action. The state-action pairs for every transition are labeled as safe or unsafe and then appended to the replay buffer. The CAM is updated every time a certain number of transitions are collected. The main differences between our training approach and the standard RL methods are three-fold: \textbf{(i)} instead of using an actor-network, the agent chooses the action among the sampled actions based on the admissibility score generated by CAM. \textbf{(ii)} the label is binary for each state-action pair instead of calculating the cumulative future rewards. \textbf{(iii)} the training objective is non-bootstrapped, which avoids the overestimation issue and is more stable. We design these three improvements around the underlying structure of the admissibility problem. We provide further explanations as follows.

\begin{algorithm}[h!]
\caption{CAM Algorithm for Training}
\label{alg::training}
\begin{algorithmic}[1]
\State \textbf{Input}: Preference function $\omega$, exploration noise $\mathcal{N}$
\NoThen
\State Initialize CAM $\phi$
\State Initialize replay buffer $R$
\label{algLine::alg1::sample_boundary}
\For{episode = 1, M} 
\For{t = 1, T}
\State Sample candidate actions $\{a\}\subseteq \mathcal{A}$
\State $\{G\}, \{\mathcal{T}\}\gets$ all agents' current states and goal states
\State Select action for each agent according to $\phi(G,a)$, $\omega(G,\mathcal{T},a)$, and $\mathcal{N}$
\State Execute the selected actions and observe the next states $s_{t+1}$
\State Label the transition ($s_t$, $a_t$, $s_{t+1}$) with $y_t$ by checking whether $s_{t+1}$ is in danger set
\State Store transition $(s_t, a_t, y_t, s_{t+1})$ in $R$
\EndFor
\State Relabel $R$ through backpropagation
\State Update $\phi$ using Equation \ref{objective}
\EndFor
\end{algorithmic}
\end{algorithm}

\noindent{\bf Sparse Reward Setting.} We model the problem as the standard Markov Decision Process (MDP) $\mathcal{M}=(\mathcal{S},\mathcal{A},T,r,\gamma)$~\cite{sutton2018reinforcement}. $S$ is the state space represented as graphs $G$. The action space $\mathcal{A}$ is continuous. The reward $r$ is sparse: it only returns non-zero values when the agent enters the region to avoid and the region to reach. These certain regions can also be dynamic, i.e., danger regions centering around other agents.

\textcolor{black}{\noindent{\textbf{Preference Function.}} Given current state $G$ and goal state $\mathcal{T}$, we assume a preference function over actions, i.e., $\omega(G,\mathcal{T},\cdot)$, is given. The preference function solely focuses on the goal-reaching. For instance, it can simply be the L2 distance from the next state to the goal, since CAM will achieve the obstacle avoidance. In the experiments, we try to choose the preference functions as simple as possible. We refer readers to the Appendix for more details.}

\noindent{\bf Action Selection via Sampling.}\label{actionselection} At every timestep, the agent chooses an action among a batch of actions sampled from the action space. Given a state $x$, the admissibility value every action $a$ is computed with $\phi(x,a)$. If there exist one or multiple actions that satisfy $\{a\mid \phi(x,a)\geq 0\}$, given the goal state $\mathcal{T}$ and a preference $\omega(x, \mathcal{T}, \cdot)$ over actions, we choose the action that has the highest preference score in the admissible set. Namely, $\argmaxA_{a}\{\omega(x, \mathcal{T}, a)|\phi(x,a)\geq 0\}$. If there does not exist any admissible action, in order to fulfill the admissibility property maximally, the agent chooses the action that has the highest admissibility score, $\argmaxA_{a}\{\phi(x,a)\}+\mathcal{N}$, where $\mathcal{N}$ is an exploration noise. We model the exploration noise under the uniform distribution in our experiment.

\noindent{\bf Training Loss of CAM.} Given admissible transitions $R_0\subseteq \mathcal{S}\times\mathcal{A}$ and inadmissible transitions $R_{d}\subseteq \mathcal{S}\times\mathcal{A}$, we train the CAM with the following objective:
\begin{multline}
    L_{B} = \frac{1}{|R_{0}|} \sum_{(x,a) \in R_{0}} \text{ReLU}(\gamma_1-\phi(x,a)) 
    + \frac{1}{|R_{d}|} \sum_{(x,a) \in R_{d}} \text{ReLU}(\gamma_2+\phi(x,a))\\
        + \frac{1}{|R_{0}|} \sum_{(x,a) \in R_{0}} \text{ReLU}(\gamma_3- \dot{\phi}(x,a) - \lambda \phi(x,a))
\label{objective}
\end{multline}

where $\gamma_1, \gamma_2, \gamma_3\geq 0$ are the coefficients to enlarge the margin of the learned boundary. Instead of simply classifying the transitions with the first and the second loss terms, we encourage the forward invariance property by adding the third term, similar to Control Lyapunov Functions and Control Barrier Functions~\cite{ames2019control}. In practice, we approximate $\dot{\phi}(x,a)$ by $\phi(x',a') - \phi(x,a)$, where $(x', a')$ is the next state-action pair on the trajectory. Intuitively, \textcolor{black}{once all pairs of consecutive transitions satisfy this condition, the transitions would form a forward invariant set, and any trajectory starting from inside the invariant set will never cross the admissible boundary. The $\lambda\phi(x,a)$ term} encourages the trajectory to be asymptotically admissible, even if the agent enters the inadmissible region sometimes.

\noindent{\bf Relabeling Through Backpropagation.}\label{method:relabel} We first label the collected transitions as admissible and inadmissible based on whether the next state is in the danger region. Once the whole trajectory terminates, we relabel these transitions through backpropagation. Intuitively, suppose the agent encounters the collision along the trajectory. In that case, it means that starting from some timestep, it enters the inadmissible region and fails to get back to the admissible set (similar to Intermediate Value Theorem~\cite{rudin1976principles}). Without relabeling, the labels of the transitions from this step to the collision event will be admissible, which is incorrect. To solve this issue, we relabel these critical transitions. 

A state-action pair $(x, a)$ on the trajectory is relabelled as inadmissible, if \textbf{(i)} the next state-action pair $(x', a')$ is inadmissible, and \textbf{(ii)} no admissible action is found at state $x'$ - namely, $\max_{a'} \phi(x',a') < 0$. Intuitively, the relabelling propagates the inadmissibility from where the agent enters the danger region back to a state, where the agent can make a decision leading to admissible regions. The first condition ensures the optimistic behavior of the CAM. We only relabel those transitions which stay in the inadmissible region and encounter danger in the future. Such optimism is important when no feasible trajectory exists at the early training stage. It encourages the agent to explore instead of predicting almost every transition to be inadmissible. The second condition ensures that there exist no admissible actions approximately by inspecting the admissibility score for all the actions. Furthermore, for the transitions that have potential admissible actions, they will not be relabelled as inadmissible easily at the early training stage since $\max_{a'} \phi(x', a') < 0$ is hard to meet for a moderately initialized CAM.

\subsection{Online Inference with Graph Decomposition}\label{method:decomposition}

At inference time, the agent behaves in the same manner as the training stage, except that there is no exploration noise during inference. When the CAM is applied directly to a large swarm of agents, however, we find that the performance degrades significantly due to the distribution shift. Such degradation occurs considerably when the inference task has a much higher density of agents. To solve this issue, we leverage the compositional property of CAMs and propose graph decomposition.

\noindent{\bf Decomposition on Large Graphs.} We can break down the whole graph $G$ into subgraphs $\{G_k\}$, where these subgraphs satisfy the training distribution, e.g., the number of agent vertices does not exceed a certain value. We construct the subgraphs repeatedly until all the edges in $G$ appear at least once in the subgraphs. For each subgraph $G_k$, and action $a$, we calculate the admissibility score as $\phi(G_k, a)$. Though each subgraph defines a less strict subtask for the agent to perform, the admissible set for the original task should lie in the intersection of all these admissible sets of the subtasks. As a result, these values on the subgraphs can be composed to represent the admissibility score on the entire graph $\phi(G, a)$, which could be totally out of distribution. To compose the admissibility values, we follow Proposition \ref{composition} and use $\phi(G, a)= \min_{G_k\subseteq G}\{\phi(G_k, a)\}$.

\begin{algorithm}[h!]
\caption{CAM Algorithm for Inference}
\label{alg::inference}
\begin{algorithmic}[1]
\NoThen
\State \textbf{Input}: CAM $\phi$, preference function $\omega$, distribution $P$ of training data
\For{t = 1, T}
\For{each agent's current state $G$, goal state $\mathcal{T}$}
\State Sample candidate actions $\{a\}\subseteq \mathcal{A}$
\State Initialize $\phi(G,a)$ as $\infty$ for all sampled candidates
\Repeat
\State Sample $G_i\subseteq G$, where $G_i\sim P$
\State $\phi(G,a)=\min(\phi(G,a), \phi(G_i,a))$ for all candidates $a$\Until{all edges on $G$ are sampled at least once}
\State Select action for the agent according to $\phi(G,a)$ and $\omega(G,\mathcal{T},a)$
\EndFor
\State Execute the selected actions and observe the next states $s_{t+1}$
\EndFor
\end{algorithmic}
\end{algorithm}

\section{Experiments}\label{Experiments}

\input{experiment}

\section{Limitations and Failure Modes} 

While the framework is generalizable in terms of scalability, it has limitations. Since the proposed GNN architecture is entirely decentralized, it disables the agents from communicating and coordinating with each other in the swarm. As a result, it could be challenging for agents to identify and avoid critical situations requiring interactions among agents, e.g., dead-locks and states with very few feasible actions. As shown in Figure~\ref{fig:overall_performance}, there is a slight drop in the safety rate of the Drone environment when the number of agents increases to 512. To understand such a failure mode, we inspect these trajectories and find that the corresponding CAM values are all negative at several consecutive preceding states right before the collision. Such behavior indicates a very low probability of finding feasible actions in these states, which validates the necessity for inter-agent communication or global coordination when the density of agents is very high in local regions. We believe that a preference function considering the global coordination can be helpful for this issue.

\section{Conclusion}

We proposed new learning-based control methods for multi-agent navigation problems by shifting the focus from training optimal control policies to training CAMs that implicitly represent a set of feasible actions. \textcolor{black}{We avoid the reward engineering by decoupling the multi-agent navigation tasks using a simple goal-reaching preference function and a learnable CAM for collision avoidance.} Our methods use GNNs to represent the CAM which takes egocentric graphs as inputs and proposes the relabelling with backtracing to learn from rich information even given sparse reward.  
In online inference, we propose the graph decomposition to deal with the covariate shift, which achieves a notably high success rate and low collision rate. While unconventional compared to the types of models routinely used in multi-agent reinforcement learning, we demonstrate that CAM has several useful properties and is particularly effective in multi-agent settings that require compositionality and generalization for scaling at inference time. In future work, we will study how to incorporate multi-agent communication and test its effectiveness in real-world scenarios. 

\acknowledgments

We thank the anonymous reviewers for their helpful comments in revising the paper. This material is based on work supported by the United States Air Force and DARPA under Contract No. FA8750-18-C-0092, AFOSR YIP FA9550-19-1-0041, NSF Career CCF 2047034, and Amazon Research Award.

\bibliography{references}

\ifarXiv
    \foreach \x in {1,...,\numbersupplementpages}
    {
        \clearpage
        \includepdf[pages={\x}]{\supplementfilename}
    }
\fi

\end{document}

\title{Appendix to “Learning Control Admissibility Models with Graph Neural Networks for Multi-Agent Navigation”}

\author{Chenning Yu\\UCSD\\chy010@ucsd.edu\And Hongzhan Yu\\UCSD\\hoy021@ucsd.edu\And Sicun Gao\\UCSD\\sicung@ucsd.edu}

\maketitle

\section{Illustration for GNN Architecture}

\begin{figure*}[h]
\centering
\includegraphics[width=\textwidth]{images/GNN_architecture.pdf}
\caption{The GNN architecture of the proposed CAM. A state for each agent is an egocentric graph $G$, which includes the node features and edge features. The CAM first encodes the graph using the GNN layer. The GNN layer then outputs the hidden state $h$ for the current agent based on the updated embedding of the corresponding node. Given action $a$ and another MLP $f$, the CAM generates the admissibility score using $\phi(G,a)=f(h,a)$.}
\label{architecture}
\end{figure*}
% In the Appendix, we first explain the details for our environments, UR5, Car, and Drone, in Section \ref{appendix:ur5}, \ref{appendix:car}, \ref{appendix:drone} respectively.

\section{Details for the UR5 Environment}\label{appendix:ur5}

\noindent{\bf Dynamics.} Each UR5 agent controls a 5D robot arm. The state $x=[\theta_1, \theta_2, \theta_3, \theta_4, \theta_5] \in [-2\pi, 2\pi]^{5}$, representing the angles on the joints of base to shoulder, shoulder to lift, elbow, wrist 1, and wrist 2. The action is the angular velocity $a=[\dot\theta_1, \dot\theta_2, \dot\theta_3, \dot\theta_4, \dot\theta_5]\in[-1,1]^5$, and the dynamics to compute the next state $x'$ are 
\begin{equation}
x' = f(x,a)=x + a \cdot dt
\end{equation}

We choose $dt=0.1$ in our experiment.

\noindent{\bf Graph Representation.} Each node $v\in V$ is simply a zero scalar. For each agent. Each edge $e:(v_i,v_j) \in E$ is a 18D vector, which concatenates 4 vectors. The first 4D vector is a one-hot vector representing the arm ID of $v_i$. The second 5D vector represents the state of the arm $v_i$. The third and fourth vectors follow the same representation as to the first and the second vectors, but those of arm $v_j$ replace the values. The graph is complete for this environment, i.e., each arm can affect any other arm in the environment.

\noindent{\bf Training Distribution.} The training distribution is that only two arms should be in the graph.

\noindent{\bf Collision and Goal-reaching Criteria.} We define collision as the contact or penetration between one arm and another arm or static obstacle. If the state of the arm is within 0.3 distance of the goal, then the goal is reached.

\noindent{\bf Preference Function $\omega$.} The preference function calculates the negative of distance to the goal based on the next state: $\omega(G,\mathcal{T},a)=-||f(x_G,a)-\mathcal{T}||^2_2$.

\noindent{\bf Hyperparameters.} For each time step, we sample $N=2000$ candidate actions for every agent. The initial learning rate of the ADAM optimizer is 3e-4, and we decrease the learning rate with a factor of 0.5 when the performance in the validation environment meets a plateau until we reach a minimum learning rate of 1e-4. We set $\gamma_1, \gamma_2, \gamma_3$ as 0, 1e-1, 1e-2. We use a batch size of 256 and update the network every 20 trajectories.

\section{Details for the Car Environment}\label{appendix:car}

\noindent{\bf Dynamics.} Each Car agent follows the Dubins' Car dynamics~\cite{dubinscar}. The state $x=[p_x, p_y,\theta], p_x,p_y\in \mathbb{R}, \theta\in[0, 2\pi)$, representing the positions on 2D plane and the heading angle. We apply a constant $v=0.05$ as the velocity for each agent. The action is the angular velocity $a=[\dot\theta]\in[-\frac{2}{3}\pi,\frac{2}{3}\pi]$, and the dynamics to compute the next state $x'$ are 

\begin{equation}
x' = f(x,a)=x + \begin{bmatrix}
v\cdot\sin(\theta+\dot\theta\cdot dt) \\
v\cdot\cos(\theta+\dot\theta\cdot dt) \\
\dot\theta\cdot dt \\
\end{bmatrix}
\end{equation}

We choose $dt=1$ in our experiment.

\noindent{\bf Graph Representation.} Each node $v\in V$ is a one-hot vector indicating whether the current node is an agent or a static obstacle. Given each edge $e:(v_i,v_j) \in E$, the feature vector of the edge is $[I, \sin(\theta_i), \cos(\theta_i), \sin(\theta_j), \cos(\theta_j), p_{x,i}-p_{x,j}, p_{y,i}-p_{y,j}]$, where $I$ is a one-hot vector, representing whether the current edge is obstacle-to-agent or agent-to-agent. For the obstacle-to-agent edges, $\sin(\theta_i), \cos(\theta_i)$ are padded as 0.

We define each node's neighbors, including agents and obstacles, as those nodes having a distance within 1.5 to the current node. We only connect edges from these neighbor nodes to each node.

\noindent{\bf Training Distribution.} The training distribution is that only 0-2 neighbor agents and 0-9 neighbor obstacles should be in the graph.

\noindent{\bf Collision and Goal-reaching Criteria.} We define the agent and the obstacle as circles with a radius of 0.15. The goals are circles with a radius of 0.3. The agent is in a collision if it is within a distance of 0.3 to other agents or obstacles. If the agent's position is within a distance of 0.45 to the 2D goal, then the goal is reached.

\noindent{\bf Preference Function $\omega$.} The preference function calculates the negative of distance to the goal based on the next state: $\omega(G,\mathcal{T},a)=-||(p_x',p_y')-\mathcal{T}||^2_2, p_x', p_y' \in f(x_G,a)$.

\noindent{\bf Hyperparameters.} For each time step, we sample $N=2000$ candidate actions for every agent. The initial learning rate of the ADAM optimizer is 1e-3, and we decrease the learning rate with a factor of 0.5 when the performance in the validation environment meets a plateau until we reach a minimum learning rate of 1e-5. We set $\gamma_1, \gamma_2, \gamma_3$ as 0, 2e-2, 1e-2. We use a batch size of 256 and update the network every 10 trajectories.

\section{Details for the Dynamic Dubins Environment}

\noindent{\bf Dynamics.}  The agent follows dynamic Dubins' car and controls a 2D action. The state $x=[p_x,p_y,v,\theta],p_x,p_y\in\mathbb{R},v\in[0,1],\theta\in[0,2\pi)$, representing the position, velocity, and heading angle in the 2D space. The 2D action $a\in\mathcal{A}=[-1,1]^{2}$ is composed of the acceleration rate $q$ and the angular velocity $\dot \theta$, and the dynamics to compute the next state $x'$ is

\begin{equation}
\begin{split}
&x' = f(x,a)=x + \begin{bmatrix}
v\cdot \sin(\theta+\dot\theta\cdot dt) \\
v\cdot \cos(\theta+\dot\theta\cdot dt) \\
q\cdot dt\\
\dot\theta\cdot dt\\
\end{bmatrix}
\end{split}
\end{equation}

We choose $dt=0.05$ in our experiment.

\noindent{\bf Graph Representation.} Each node is a one-hot vector indicating whether the current node is an agent or a static obstacle. Given each edge $e:(i,j) \in E$, the feature vector of the edge is $[I, v_i, v_j, \sin(\theta_i), \cos(\theta_i), \sin(\theta_j), \cos(\theta_j), p_{x,i}-p_{x,j}, p_{y,i}-p_{y,j}]$, where $I$ is a one-hot vector, representing whether the current edge is obstacle-to-agent or agent-to-agent. For the obstacle-to-agent edges, $v_i,\sin(\theta_i), \cos(\theta_i)$ are padded as 0.

We define each node's neighbors, including agents and obstacles, as those nodes having a distance within 1.5 to the current node. We only connect edges from these neighbor nodes to each node.

\noindent{\bf Training Distribution.} The training distribution is that only 0-2 neighbor agents and 0-9 neighbor obstacles should be in the graph.

\noindent{\bf Collision and Goal-reaching Criteria.} We define the agent and the obstacle as circles with a radius of 0.15. The goals are circles with a radius of 0.3. The agent is in a collision if it is within a distance of 0.3 to other agents or obstacles. If the agent's position is within a distance of 0.45 to the 2D goal, then the goal is reached.

\noindent{\bf Preference Function $\omega$.} The preference function calculates the negative of distance to the goal based on the next state: $\omega(G,\mathcal{T},a)=-||(p_x',p_y')-\mathcal{T}||^2_2, p_x', p_y' \in f(x_G,a)$.

\noindent{\bf Hyperparameters.} For each time step, we sample $N=2000$ candidate actions for every agent. The initial learning rate of the ADAM optimizer is 1e-3, and we decrease the learning rate with a factor of 0.5 when the performance in the validation environment meets a plateau until we reach a minimum learning rate of 1e-5. We set $\gamma_1, \gamma_2, \gamma_3$ as 0, 2e-2, 1e-2. We use a batch size of 256 and update the network every 10 trajectories.

\section{Details for the Drone Environment}\label{appendix:drone}

\noindent{\bf Dynamics.}  Each Drone agent follows a 3D quadrotor model adopted from~\cite{dawson2022safe} and controls a 4D action. The state $x=[p_x, p_y,p_z,v_x,v_y,v_z,\alpha,\beta,\gamma], p_x,p_y,p_z\in \mathbb{R}, v_x,v_y,v_z\in[-1, 1], \alpha,\beta,\gamma\in[-\pi/2,\pi/2]$, representing the position and velocity in the 3D space, and the angles in pitch, roll, and yaw. The action is the instant acceleration $q$ and the angular velocities, i.e.,  $a=[q, \dot\alpha,\dot\beta,\dot\gamma]\in[-1,1]^{4}$, and the dynamics to compute the next state $x'$ is

\begin{equation}
\begin{split}
&f(x, t+dt) = f(x, t) + \begin{bmatrix}
v_x \\
v_y \\
v_z \\
-\sin(\beta)\cdot q \\
\cos(\beta)\sin(\alpha)\cdot q \\
\cos(\beta)\cos(\alpha)\cdot q -g \\
\dot\alpha \\
\dot\beta \\
\dot\gamma \\
\end{bmatrix}\cdot dt\\
&f(x, 0) = x\\
&x' = f(x, k \cdot dt)\\
\end{split}
\end{equation}

$g$ is the gravity acceleration of the Earth as 9.8. We choose $dt=0.01$ and $k=10$ in our experiment.

\noindent{\bf Graph Representation.} Each node $v\in V$ is a one-hot vector indicating whether the current node is an agent or a static obstacle. Given each edge $e:(v_i,v_j) \in E$, the feature vector of the edge is $[I, v_{x,i}, v_{y,i}, v_{z,i}, \sin(\alpha_i), \cos(\alpha_i), \sin(\beta_i), \cos(\beta_i), v_{x,j}, v_{y,j},$ $v_{z,j}, \sin(\alpha_j), \cos(\alpha_j), \sin(\beta_j), \cos(\beta_j), p_{x,i}-p_{x,j}, p_{y,i}-p_{y,j}, p_{z,i}-p_{z,j}]$, where $I$ is a one-hot vector, representing whether the current edge is obstacle-to-agent or agent-to-agent. For the obstacle-to-agent edges, $v_{x,i}, v_{y,i}, v_{z,i}, \sin(\alpha_i), \cos(\alpha_i), \sin(\beta_i), \cos(\beta_i), p_{z,i}-p_{z,j}$ are padded as 0.

We define each node's neighbors, including agents and obstacles, as those nodes having a distance within 1.5 to the current node. We only connect edges from these neighbor nodes to each node.

\noindent{\bf Training Distribution.} The training distribution is that only 0-2 neighbor agents and 0-9 neighbor obstacles should be in the graph.

\noindent{\bf Collision and Goal-reaching Criteria.} We define the agent as spheres with a radius of 0.15. The obstacles are vertical cylinders with a radius of 0.15 and infinite height. The goals are spheres with a radius of 0.3. The agent is in a collision if it has a distance within 0.3 to other agents or obstacles in the 3D space. An agent reaches its goal if its 3D position has a distance of 0.45 to its goal.

\noindent{\bf Preference Function $\omega$.} The preference function calculates the negative of an error between the action $a$ and a linear–quadratic regulator (LQR) controller $\pi$. The LQR controller $\pi$ takes the goal and the current state, and gives an action which only considers the goal-reaching: $\omega(G,\mathcal{T},a)=-||\pi(x_G,\mathcal{T})-a||^2_2$.

\noindent{\bf Hyperparameters.} For each time step, we sample $N=2000$ candidate actions for every agent. The initial learning rate of the ADAM optimizer is 1e-3, and we decrease the learning rate with a factor of 0.5 when the performance in the validation environment meets a plateau until we reach a minimum learning rate of 1e-5. We set $\gamma_1, \gamma_2, \gamma_3$ as 0, 1e-1, 1e-2. We use a batch size of 256 and update the network every 20 trajectories.

\section{Details for the Chasing Game}

We provide more details for the chasing game in this section. We define the safety rates of the chasing game in the same way as those of the navigation task. The reward is defined in another way, to reflect on the situation that one Drone agent could drift away from its target agent though it executes the most preferred action. We want to avoid penalizing such behavior because the most preferred action should not be discouraged if it is also safe at the same time. We define reward as follows: $ R=\sum_{t=1}^T \text{clip}(d_{t-1} - d_{t}, 0, 2)$, where $d_{t-1}$ and $d_{t}$ indicate the distances between the agent and the target agent in 2D or 3D space at $t-1$-th and $t$-th step. Such a reward keeps track of the agent's improvement at every step while not penalizing on no progress. The preference for the agent follows the same way we defined in the navigation tasks of the Car and Drone. We update the goal for each agent to be the position of the target agent at every step. 

\section{Details for the Experiments}

\noindent{\bf Reproducing the Baselines.} The inputs to the networks of DDPG and MACBF are the same as those to the CAM. For DDPG, we use the GNN architectures as the actor and Q-critic network. The GNN actor takes an egocentric graph for each agent and outputs the action. The GNN critic takes the graph and action and outputs the Q value. For MACBF, since it requires differentiable dynamics, we implement an additional dynamic function using PyTorch to guarantee the differentiability. When computing the derivative of the CBF value over time, we assume the graph structure remains the same, i.e., no adding or removing edges. We also find that replacing the temporary dataset of MACBF with the replay buffer, typically used for the RL methods, improves the stability of MACBF.

\textcolor{black}{\emph{Notes on why MA-DDPG is not deployed here \cite{maddpg}}: Our GNN implementation deviates from the MA-DDPG algorithm because of scalability concerns. In MA-DDPG, the actor network for each agent is independent and does not share the weight. As a result, MA-DDPG requires the same number of agents for the training and inference tasks. However, in our experiment, since we focus on the generalization for scalability, the number of agents during inference varies, while our training tasks are fixed to have three agents. Thus, it is impossible to deploy the MA-DDPG algorithm.}

\noindent{\bf Running the Experiments.} All the methods are trained on CPU as Intel Core i7-11700F. The GPU is NVIDIA RTX 3080. We train each method until we reach convergence, which typically requires fewer than two days.

\noindent{\bf Accelerating the Computation.}\label{sec:acceleration} Though the GNN can be deployed on decentralized devices at inference time, we conduct our experiment on one single desktop, which requires computation optimization. Assume at a time step of the inference time that there are 512 agents in the environment. We need to test 2000 actions for each agent, and each agent samples at least 100 subgraphs. We can infer that there are at least $100\times2000\times512\approx 1.02\times 10^8$ forward operations for the GNN to compute. We need to accelerate such computation; otherwise, the computational cost will be intractable.

We conduct three improvements for the acceleration. The first improvement is to merge the computation of the hidden state for all state-action pairs which share the same input graph into one forward passing in the GNN layer, as mentioned in Section \ref{alg::gnn_for_cam}. The second improvement is to parallelize the computation of the CAM value by stacking the pairs of hidden states and actions as a matrix and then feeding it to the GPU. Such stacking enables the computation with batches of state-action pairs. The third improvement is to choose which agents to compute for each batch adaptively. By inspecting the CAM values of the computed batches, if we have already found admissible actions for one specific agent, then the rest of the state-action pairs for this agent do not need to be computed, and the agent chooses the action among those admissible actions that have been computed. By filtering out the determined agents, we save the unnecessary computation of the state-action pairs of these agents.

\section{Ablation Study: Varying Obstacle Density}
\begin{figure*}[h]
\centering
\includegraphics[width=0.7\textwidth]{images/overall_performance_obstacle.pdf}
\caption{The performance of all methods on Car and Drone with regard to varying obstacle numbers.}\label{fig:overall_obstacle}
\end{figure*}  

In the main paper, we test the scenarios where the densities of both the agents and the obstacles vary. In this section, we focus on how the obstacle density affects the performance of each method.

We fixed the map size to be 4x4 and increased the obstacle number from 0 to 20. We average the results over 100 randomly generated environments. We do not include the UR5 environment since the obstacle number for UR5 is fixed to be 4. As shown, the CAM significantly outperforms other methods and maintains the safety rate near 100\%. On the other hand, MACBF and DDPG show a lack of robustness when the obstacle density increases. We also observe a remarkable degradation of the performance of MACBF for the Drone environment. When there are 0 obstacles, the safety rate of MACBF is close to 99\%, but then the safety rate jumps to around 90\% when there are 5 obstacles. Such a result shows that MACBF works better to avoid inter-agent collisions and cannot deal well with obstacle-agent collision avoidance. This result explains why the performance of MACBF improves while the density of obstacles decreases in the Drone environment in Figure \ref{fig:overall_performance}.

\section{\textcolor{black}{Inference Time}}

Figure \ref{fig:running_time} illustrates the running time during inference. We average the running time over all the agents and all the time steps in the test cases. With the three acceleration improvements mentioned in Section \ref{sec:acceleration}, we can now reduce the running time to around 4.3 milliseconds at most. The computation time can further be improved when deployed in real-world experiments. In that case, each agent can compute the admissible set independently, since our GNN architecture is fully decentralized. It only requires observations of nearby agents and obstacles and does not need inter-agent communication.

We find that in Figure \ref{fig:running_time}, the trends of the running time for Car and Drone are similar - the running time decreases and then increases as the density of agents grows. The running time first decreases, since our method utilizes the GPU and computes the admissible actions of all the agents in parallel. However, when the number of agents grows to a certain threshold between 32 and 128, the memory of the GPU will reach its limitation and cannot take more agents beyond this threshold. If the number of agents is higher than this threshold, we need multiple GPU forward passes to cover all the agents. As a result, the running time increases afterward. More GPU memory should make the computation even faster in such a centralized computation scenario.

\begin{figure*}[h]
\centering
\includegraphics[width=0.7\textwidth]{images/running_time.pdf}
\caption{The average running time of our method on Car and Drone during inference for each agent at each time step (in milliseconds). The running time is averaged over the 100 test cases.}\label{fig:running_time}
\end{figure*}  

\section{\textcolor{black}{Details for the Single Agent Example}}

\begin{figure*}[h]
\centering
\includegraphics[width=\textwidth]{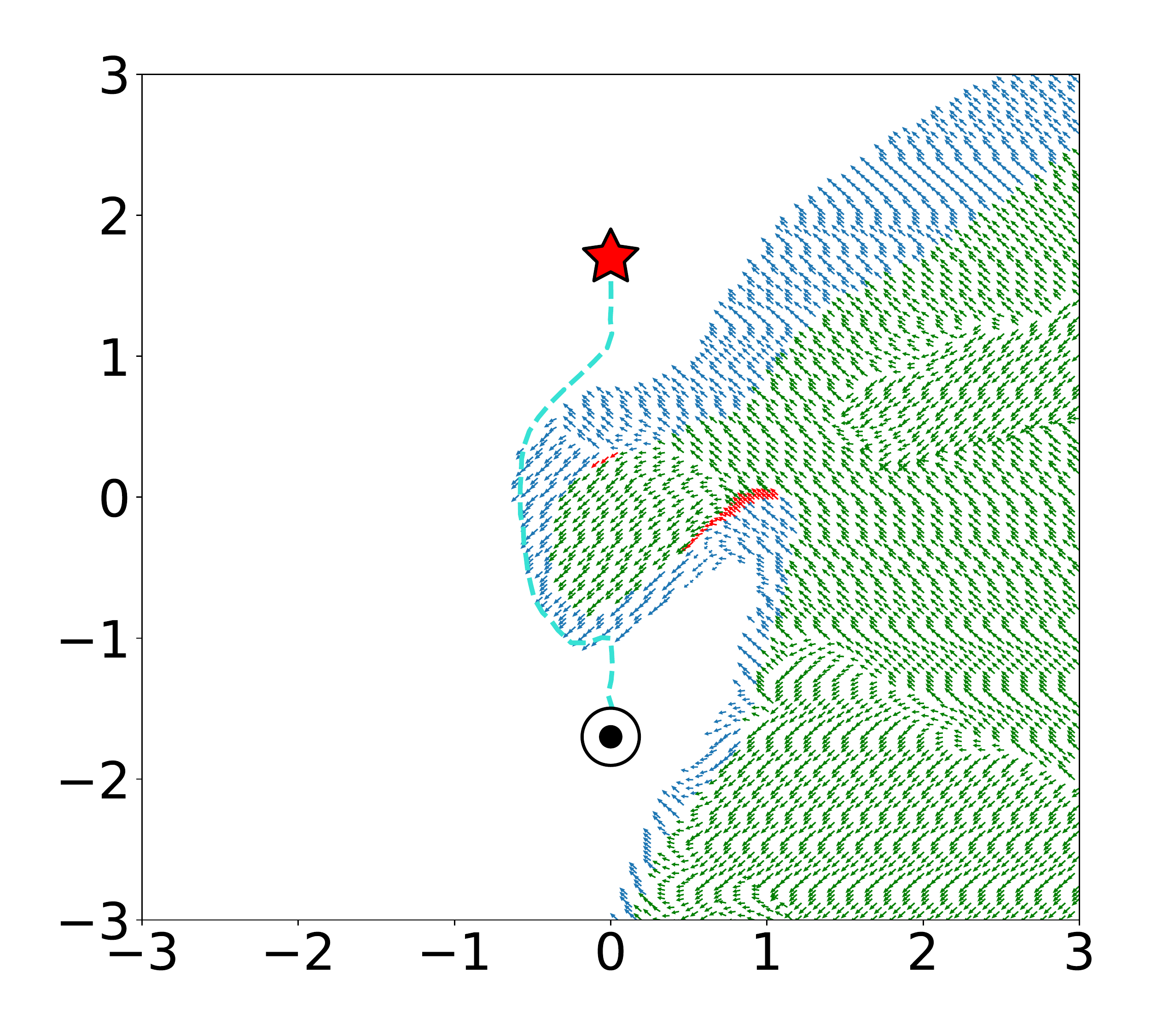}
\caption{We show that the forward-invariance property holds along the trajectory. The direction of each arrow indicates the most admissible action to take on each state $x$: $\arg\max_{a\in\mathcal{A}}\phi(x,a)$. The green arrows represent the states in the inadmissible region $\{x\mid\forall a\in\mathcal{A}, \phi(x,a)<0\}$. Both the blue arrows and the red arrows represent the states on the boundary: $\{x\mid \exists a_1, a_2\in\mathcal{A}, \phi(x,a_1)\geq 0, \phi(x,a_2)<0\}$. The blue arrows are those boundary states that hold the forward-invariance property. The red arrows are those boundary states that enter the inadmissible region after executing the most admissible action. Note: Since we only train on transitions from collected trajectories, CAM can be asymmetric and violate the forward-invariance property at some unseen states. }\label{fig:arrow_boundary}
\end{figure*}  

In this section, we provide an additional single-agent experiment as a proof of concept. This example has more complex dynamics than the one in the main paper. We provide the details as follows.

\noindent{\bf Dynamics.}  The agent here follows the single integrator and controls a 2D action. The state $x=[p_x,p_y]$, representing the position in the 2D space. The 2D action $a=[v_x,v_y]\in\mathcal{A}=[-1,1]^{2}$ is composed of the velocities along x and y axis, and the dynamics to compute the next state $x'$ is

\begin{equation}
\begin{split}
&x' = f(x,a)=x + \begin{bmatrix}
v_x\\
v_y\\
\end{bmatrix}\cdot dt
\end{split}
\end{equation}

We choose $dt=0.05$ in our experiment. In this example, we use an MLP to represent CAM, since the state here is solely a vector.

\noindent{\bf Task Distribution.} The goal state $[p_x=0, p_y=1.7]$, and the obstacles are all fixed during training and test. The starting state of the agent is fixed as $[p_x=0, p_y=-1.7]$.

\noindent{\bf Collision and Goal-reaching Criteria.} We define the agent as a circle with a radius of 0.15. The goals are circles with a radius of 0.15. The agent is in a collision if it intersects with the obstacles. If the agent intersects with the 2D goal, then the goal is reached.

\noindent{\bf Preference Function $\omega$.} The preference function calculates the negative of distance to the goal based on the next state: $\omega(x,\mathcal{T},a)=-||(p_x',p_y')-\mathcal{T}||^2_2, [p_x', p_y'] = x'=f(x,a)$.

\subsection{Forward-Invariance Property}

To further investigate whether the learned CAM holds the forward-invariance property, we illustrate the boundary of the CAM $\{x\mid \exists a_1, a_2\in\mathcal{A}, \phi(x,a_1)\geq 0, \phi(x,a_2)<0\}$ in Figure \ref{fig:arrow_boundary}.

We show that the trajectory never leaves the admissible region of CAM, i.e., the trajectory holds the forward-invariance property. Once the trajectory meets the boundary, it starts to follow the most preferred action among admissible actions, instead of the most preferred action among all sampled actions. Meanwhile, we find that around 0.9\% states in the boundary states violate the forward-invariance property, i.e., they enter the inadmissible region after executing the most admissible action.

We emphasize that since we only train on transitions from collected trajectories, CAM can be asymmetric and violate the forward-invariance property at some unseen states. We design the CAM to maintain the forward invariance and focus on states that could exist along possible trajectories. If the agent does not meet some states at any time step during inference, then we do not require these states to obey the forward invariance. Empirically, we have shown that as long as the training tasks and inference tasks are under the same distribution, the forward invariance often holds for inference trajectories once the CAM learns from enough training data.

\section{\textcolor{black}{Additional Single Agent Environment: Dynamic Dubins}}

\begin{figure*}[h]
\centering
\includegraphics[width=1.0\textwidth]{images/dynamic_dubins_arrow.pdf}
\caption{An additional single-agent example for the proposed CAM dealing with more complex dynamics, where the agent needs to catch failure early on. The red star represents the goal state, and three blue boxes represent the obstacles. The orange region denotes the set of initial states. We illustrate state-action pairs $(x,a)$ on sampled trajectories starting from initial states. All trajectories reach the goal successfully with no collisions. The direction of each arrow indicates the action that CAM agent takes on each state $x$. The blue arrows represent the admissible region $\{x\mid\forall a\in\mathcal{A}, \phi(x,a)\geq 0\}$. Both the red and the green arrows represent the states on the boundary: $\{x\mid \exists a_1, a_2\in\mathcal{A}, \phi(x,a_1)\geq 0, \phi(x,a_2)<0\}$. Only the red arrows represent those boundary transitions that violate the forward-invariance property. The black arrows represent the states in the inadmissible region $\{x\mid\forall a\in\mathcal{A}, \phi(x,a)< 0\}$.}\label{fig:dynamic_dubins}
\end{figure*}  

In this section, we provide an additional single-agent experiment as a proof of concept. This example has more complex dynamics than the one in the main paper.

\subsection{Forward-Invariance Property}

We sample around 40 trajectories starting from the initial states, and illustrate the collected transitions in Figure \ref{fig:dynamic_dubins}. The direction of each arrow indicates the action that CAM agent takes on each state $x$. The blue arrows represent the admissible region $\{x\mid\forall a\in\mathcal{A}, \phi(x,a)\geq 0\}$. Both the red and the green arrows represent the states on the boundary: $\{x\mid \exists a_1, a_2\in\mathcal{A}, \phi(x,a_1)\geq 0, \phi(x,a_2)<0\}$. Only the red arrows represent those boundary transitions that violate the forward-invariance property. The black arrows represent the states in the inadmissible region $\{x\mid\forall a\in\mathcal{A}, \phi(x,a)< 0\}$.

We show that around 98.6\% of transitions along the trajectories are admissible and forward invariant. There are around 27.7\% of states on the boundary. Around 0.4\% of states violate the forward-invariance property, i.e., the CAM agent crosses the boundary and enters the inadmissible region after executing the action. There are 1\% of transitions that are inadmissible, but all of them return back to the admissible regions after executing sequences of most admissible actions.

\subsection{Details for the Additional Single Agent Environment}\label{single_dynamic_dubins}

\noindent{\bf Dynamics.}  The agent here is one dynamic Dubins' car and controls a 2D action. The state $x=[p_x,p_y,v,\theta],p_x,p_y\in\mathbb{R},v\in[0,1],\theta\in[0,2\pi)$, representing the position, velocity, and heading angle in the 2D space. The 2D action $a\in\mathcal{A}=[-1,1]^{2}$ is composed of the acceleration rate $q$ and the angular velocity $\dot \theta$, and the dynamics to compute the next state $x'$ is

\begin{equation}
\begin{split}
&x' = f(x,a)=x + \begin{bmatrix}
v\cdot \sin(\theta+\dot\theta\cdot dt) \\
v\cdot \cos(\theta+\dot\theta\cdot dt) \\
q\cdot dt\\
\dot\theta\cdot dt\\
\end{bmatrix}
\end{split}
\end{equation}

We choose $dt=0.05$ in our experiment. In this example, we use an MLP to represent CAM, since the state here is solely a vector.

Note that the maximum acceleration rate is saturated here. At each time step, the agent can change the velocity by increasing or decreasing 0.05 at most. As a result, the agent needs to catch failure early on.

\noindent{\bf Task Distribution.} The goal state $(p_x=0, p_y=1.7)$, and the obstacles are all fixed during training and test. The starting state of the agent is randomized within the ditribution of $p_x\in[-0.25, 0.25], p_y=-1.7,v=0, \theta=\frac{\pi}{2}$.

\noindent{\bf Collision and Goal-reaching Criteria.} We define the agent as a circle with a radius of 0.15. The goals are circles with a radius of 0.15. The agent is in a collision if it intersects with the obstacles. If the agent intersects with the 2D goal, then the goal is reached.

\noindent{\bf Preference Function $\omega$.} The preference function calculates the negative of distance to the goal based on the next state: $\omega(x,\mathcal{T},a)=-||(p_x',p_y')-\mathcal{T}||^2_2, p_x', p_y' \in f(x,a)$.

\subsection{Performance Progress in the Additional Single Agent Environment}

\begin{figure*}[!htb]
\centering
\includegraphics[width=1.0\textwidth]{images/dynamic_dubins.png}
\caption{The progress of the performance during training. The x-axis for each figure is the number of sampled trajectories during training. From left to right: \textbf{(i)} Averaged success rate over a window of the nearest 20 trajectories. A trajectory succeeds if and only if the agent reaches the goal with no collision. \textbf{(ii)} The number of relabelled transitions along each trajectory. \textbf{(iii)} The ratio of admissible actions $p(\phi(x,a)\geq 0), a\sim\mathcal{A}$, averaged over all states $\{x\}$ along each trajectory.}\label{fig:dynamic_dubins_perf}
\end{figure*}  

After the training converges, we sample the initial state from $p_x=-0.25$ to $p_x=0.25$ with an interval equal to $0.05$. As shown in Figure \ref{fig:dynamic_dubins}, all the sampled trajectories reach the goal successfully with no collisions. We provide the video in the supplementary material with more details.

Additionally, we show the progress during training in Figure \ref{fig:dynamic_dubins_perf}. We observe that a great improvement in the success rate happens right after our algorithm relabels a relatively high number of transitions. This observation is consistent with our relabelling mechanism design, which aims to catch the failure early using backtracing. 

We also observe a slight improvement starting around the 700-th trajectory, where the success rate increases gradually from 95\% to 100\%. This happens at the same time when the ratio of admissible actions drops. We infer that at that specific time, the agent has explored some new states, which are critical for leading the agent to success. Though the agent is not familiar with these states initially, by collecting these new experiences and learning from them, the ratio becomes stable again and the success rate grows afterward.

%% file: related.tex
\noindent{\bf Multi-Agent Navigation.} Classical Multi-Agent Path Finding is one of the most popular definitions of the multi-agent navigation task, where the state space lies in a pre-constructed graph shared among agents~\cite{stern2019multi}. Under this setting, heuristic-based methods have been used to generate high-quality
solutions, such as Priority-Based Search and Conflict-Based Search~\cite{pbs,cbs}. Learning-based methods have also been proposed to accelerate the planning time, using CNN and GNN \cite{qingbiao1,qingbiao2,primal}. In this work, we focus more on the continuous state space with reactive planning. Traditional methods, including Optimal Reciprocal Collision Avoidance~\cite{orca}, Buffered Voronoi Cells~\cite{bandyopadhyay2017probabilistic}, and Artificial Potential Functions~\cite{potential86,Barraquand1992APF,Ge2002APF,hernandez2011apf}, are designed manually to address specific structural properties. On the other hand, learning-based multi-agent  methods using reinforcement learning and imitation learning~\cite{macommunicate,gpg,PIC,maddpg,QMIX,glas,neuralswarm}, may or may not generalize when there is covariant shift for test tasks. Recently, MACBF~\cite{macbf} incorporates multi-agent safe certificates~\cite{ames2019control,DBLP:conf/amcc/HanHP18,wan2017semi} into a learning-based framework, which shows the generalization capability while preserving the safety properties. Nevertheless, in \cite{macbf} the learning of the neural controller requires a reference controller, which may constrain the potential of the neural controller, when the actions dissimilar to the reference actions could yield better rewards and safer results. Though learned in an online fashion, our work only assumes sparse reward and avoids the reward hacking problem.

% \citet{} uses Sum-of-Squares to approximate the sub-level set of a Lyapunov-like barrier function for multi-agent coordination. \citet{} models the multi-robot formation as an optimization problem, and gives a theoretical analysis of the formation's bias. 
% \paragraph{Certifiable Control}
 
\noindent{\bf Graph Neural Networks.} Graph Neural Networks (GNN), including Graph Convolutional Networks~\cite{GCN}, Message Passing Neural Networks~\cite{MPNN}, Interaction Networks~\cite{interactionn}, are deep neural networks that operate on graph-structured
data~\cite{gnnancestor}. Graph neural networks are permutation-equivariant to the orders of nodes on graph~\cite{deepset,pointnet}, which is important for homogeneous multi-agent problems since the order among agents should not matter. In robotics, the effectiveness of GNN has been shown on tasks such as path planning~\cite{qingbiao1,qingbiao2}, motion planning~\cite{gnnmp,chenning2021collision}, coverage and exploration~\cite{prorok1,tolstaya2020multi}, and SLAM~\cite{superglue}. In this work, we use GNN as an architecture to implement CAM due to the multi-agent navigation problem can be modeled as a graph-based problem in nature. However, we believe that the proposed CAM could generalize to other inputs that are not graphs.

%% file: experiment.tex
\subsection{Proof of Concept: CAM for Single Agent Environment}\label{singleagentenv}

\begin{figure*}[h]
\centering
\includegraphics[width=1.0\textwidth]{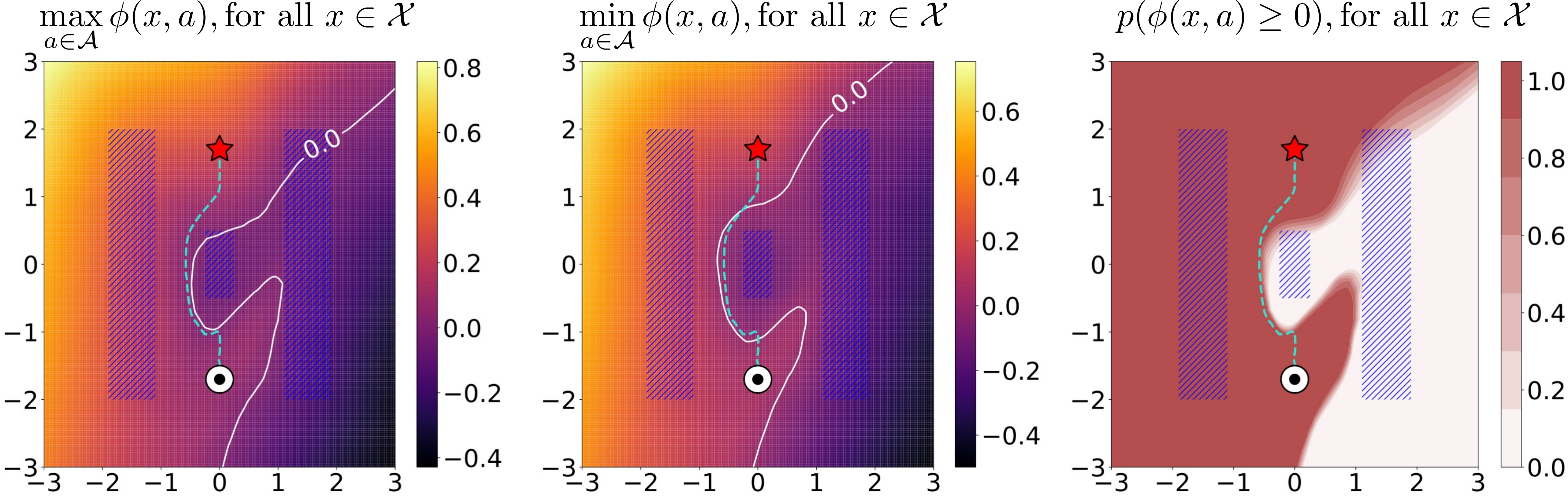}
\caption{A proof of concept for the proposed CAM in a single-agent environment. The agent aims to reach the goal while avoiding obstacles. The black point denotes the starting state, and the red star denotes the goal state. The shaded blue regions denote the obstacles. Given the learned CAM $\phi$, action space $\mathcal{A}:\{-0.1, 0.1\}^2$, and an arbitrary state $x \in \mathcal{X}:\{-3, 3\}^2$, we illustrate $\max_{a\in\mathcal{A}}\phi(x,a)$, $\min_{a\in\mathcal{A}}\phi(x,a)$, and $p(\phi(x,a)\geq 0)$ respectively. We show that the learned CAM never leaves the admissible set, i.e. $\{x\in\mathcal{X}:\max_{a\in\mathcal{A}}\phi(x,a)\geq 0\}$, along the trajectory.}
\label{fig:single_env}
\end{figure*}

We begin by analyzing the CAM agent in a single-agent environment. We perform such analysis using a 2D minimal example environment, where we place three danger regions to form narrow passages, as shown in Figure \ref{fig:single_env}. We illustrate $\max_{a\in\mathcal{A}}\phi(x,a)$, $\min_{a\in\mathcal{A}}\phi(x,a)$, and $p(\phi(x,a)\geq 0)$ respectively. Intuitively, $\max_{a\in\mathcal{A}}\phi(x,a)$ indicates whether there exists any admissible actions that drive the agent to the admissible set. $\min_{a\in\mathcal{A}}\phi(x,a)$ indicates the possibility to enter the inadmissible sets if actions are picked randomly. As what we show in the figure, the trajectory (blue dashed line) does not intersect with the admissibility boundary, i.e. $\{x\in\mathcal{X}:\max_{a\in\mathcal{A}}\phi(x,a)=0\}$. This example stays consistent with the forward-invariance objective for learning the CAM: if the agent executes an admissible action $a\in\mathcal{A}: \phi(x,a)\geq 0$ at every time step, then it will never leave the admissible set.

\subsection{Multi-agent Navigation Experiments}

\begin{figure*}[!h]
\centering
\includegraphics[width=\textwidth]{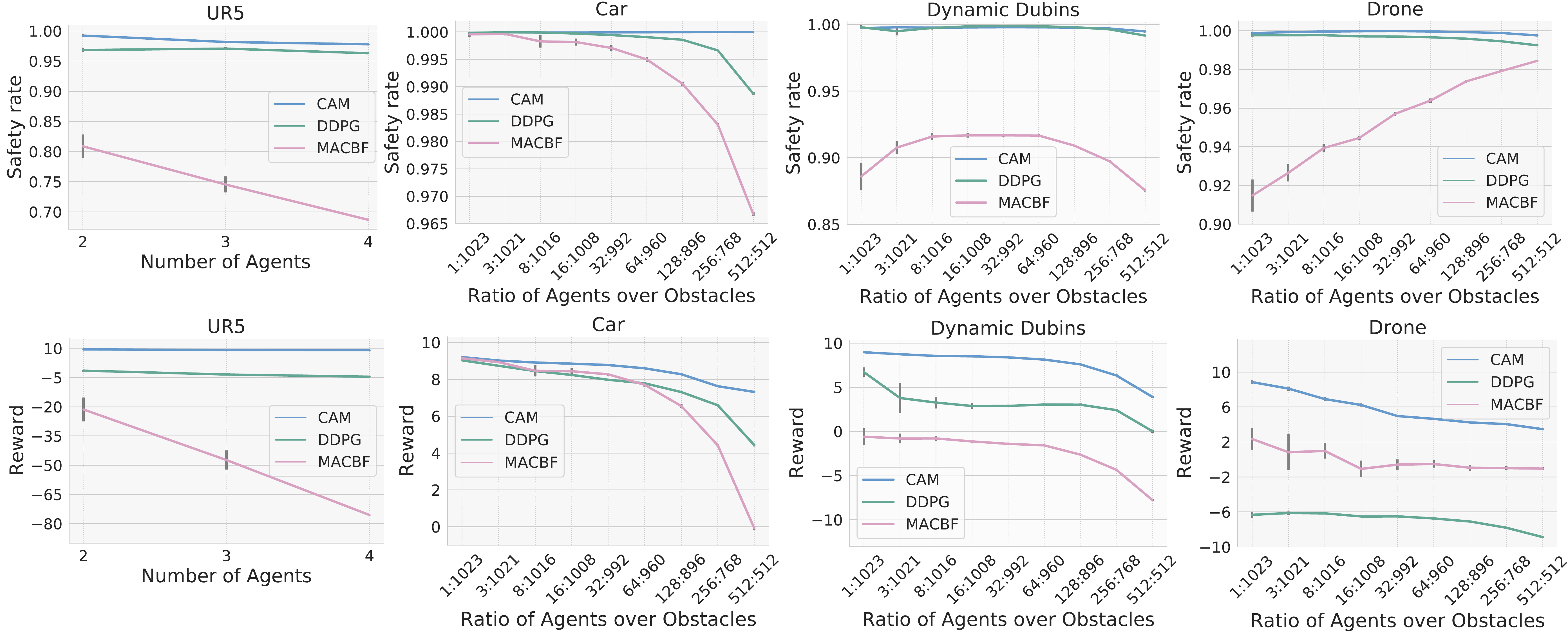}
\caption{We show the performances of CAM and prior methods in the UR5, Car, Dynamic Dubins, and Drone environment, where each method is trained with 3 agents during training but tested up to 512 agents. CAM substantially outperforms prior approaches in all the environments.}
\label{fig:overall_performance}
\vspace{-0.5cm}
\end{figure*}

\noindent{\bf Experiment Setup.} We evaluate our methods under 4 types of environments: UR5, Car, Dynamic Dubins, and Drone. We briefly describe these 4 environments as follows: \textbf{(i)} The UR5 environment is implemented with PyBullet~\cite{pybullet} and contains 4 decentralized 5D robot arms with cubic obstacles. We randomly select solely 2 robot arms from these four arms during training. We fix the configurations of the static obstacles and the goal states of the arms. In order to make the test task solvable, the initial states are randomized during training and fixed during testing. \textbf{(ii)} For the Car environment, each agent follows the 2D Dubins' car model~\cite{dubinscar}, and controls a 1D action. The static obstacles are circles with a fixed radius. \textbf{(iii)} The Dynamic Dubins agent controls a 2D action and follows a modified Dubins' Car dynamic which enables the control of acceleration. \textbf{(iv)} For the Drone environment, each agent follows a 3D quadrotor model adopted from~\cite{dawson2022safe} and controls a 4D action. Each obstacle is a cylinder with infinite height. For both Car and Drone environments, we only train on environments with 3 agents, randomized initial states and goal states, and a set of random obstacles. The map size is fixed as 3x3 during training while varying during testing. We provide full details of system dynamics and environments in the Appendix.

To evaluate our CAM approach thoroughly, we design diverse scenarios for all four environments. For the UR5 environment, we make the number of agents vary from 2 to 4. For scenarios with 2 or 3 arms, the arms are selected randomly from the four decentralized arms. For the other environments, we fix the map size as 32x32 and vary the ratio of agents over obstacles from 1:1023 to 512:512. For each setting, we average the performance over 100 randomly generated test cases.

\noindent{\bf Baselines.} The baseline approaches we compare with include DDPG~\cite{ddpg} and MACBF~\cite{macbf}. DDPG is a standard RL approach to learning safe behavior through maximizing rewards. MACBF is a state-of-the-art approach for multi-agent safe control problems, which learns a policy jointly with neural safe certificates. We reimplement all the baselines with GNNs for a fair comparison. \textcolor{black}{Note that our GNN implementation for DDPG is a multi-agent algorithm, since each agent is aware of the other agents by taking the egocentric graph as the input. We refer reader to Appendix for more details.}

\noindent{\bf Evaluation Metrics.} We evaluate our methods based on two metrics: the safe rate and the average reward. The safe rate counts the number of collisions for each agent at every time step and averages over all the agents through the whole trial~\cite{macbf}. At each time step, the reward is -1 if the agent collides with another agent or the static obstacle. A +10 reward will be given at the end of the trajectory, if the agent reaches its goal at any time step along the trajectory. A small negative constant of -0.1 is added to the reward to encourage shorter paths for goal-reaching. The reward is accumulated throughout the trial and averaged over all the agents. 

\noindent{\bf Overall Performance.} In Figure~\ref{fig:overall_performance}, we demonstrate the overall performances for the UR5, Car, Dynamic Dubins, and Drone environments. Our method shows the generalization across different environments and densities of agents and obstacles. Though trained with only 3 agents, the safety rate of CAM remains nearly 100\% for all the environments, up to 512 agents. Even if the density of agents and obstacles deviates from the training environment, our CAM approach can avoid the distribution shift, using decomposition on large graphs and the compositionality of CAMs.

On the other side, though our implementation for MACBF and DDPG can take an arbitrary number of agents using GNNs, the performance degrades significantly when the distribution of input graphs shifts in test environments. We also inspect that the averaged reward for all the methods decreases in most cases when the number of agents increases. It is due to more agents failing to reach the goal when the timeout happens since it takes more time for agents to avoid each other. We encourage the reader to view the video demonstrations in the supplementary material for all 4 environments.

\subsection{Zero-Shot Transfer to Chasing Game}\label{exp:chasing}

\begin{figure*}[!th]
\begin{minipage}[c]{0.3\textwidth}
\centering
\includegraphics[scale=0.3]{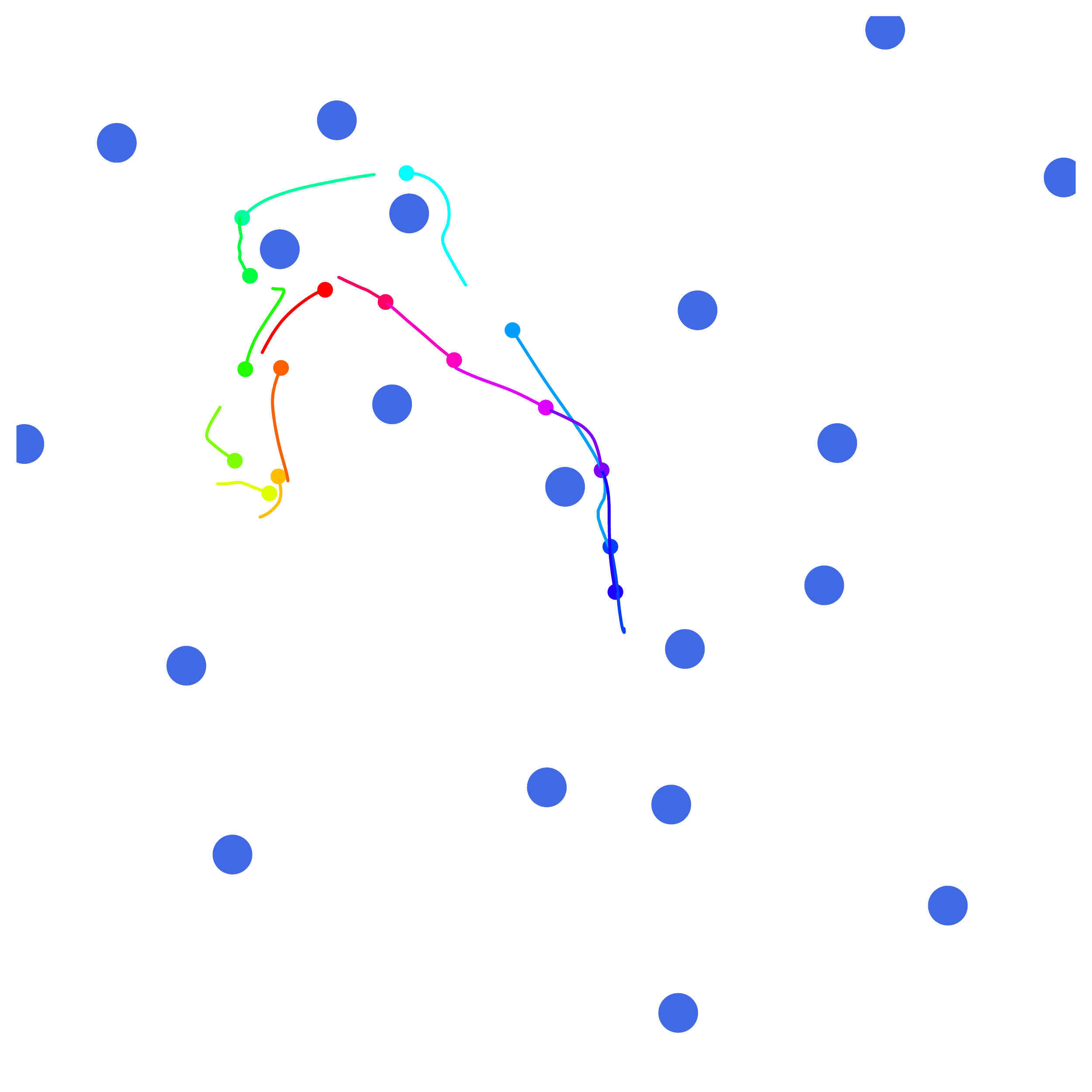}
\end{minipage}
\begin{minipage}[c]{0.7\textwidth}
\centering
\begin{tabular}{c c c}
\toprule  {\bf Method} & {\bf CAM} & {\bf CAM w/o decomposition} \\
\midrule
Safety Rate (Car) & 0.99$\pm$0.00 & 0.90$\pm$0.03\\  
\midrule
Reward (Car) &3.10$\pm$1.20 & -19.98$\pm$7.08\\
\midrule
\midrule
Safety Rate (Drone) &0.989$\pm$ 0.005 & 0.988$\pm$0.008\\  
\midrule
Reward (Drone) &6.84$\pm$1.12 & 6.54$\pm$2.15\\
\bottomrule
\end{tabular}
\end{minipage}
\captionof{table}{Left: A snapshot of a trajectory in the chasing game with 16 Drone agents from a bird's-eye view. The agents are the colorful points, and the obstacles are the blue circles of a larger size. Each agent aims to chase another agent without collision. Right: Performances of CAMs with and without graph decomposition on 64 agents, which shows the effectiveness of the graph decomposition.}
\label{table:chasing}
\end{figure*}

In this section, we study the generalization of our CAM model for zero-shot transfer to other multi-agent tasks, e.g., the chasing game. In the chasing game, each agent chases another agent to pursue. The target agent is assigned randomly at the beginning of every episode. The agent's goal is to maintain the distance to the assigned agent, with no collision with agents and obstacles. We fix the number of agents to 64 in the experiment. This new task is substantially different from the task for training, where the assigned goal is fixed. We provide more details of the setting in the Appendix.

In Table \ref{table:chasing}, we show that our CAM method can adapt to the new task effectively due to its safety and goal-reaching decoupling property, preserving a relatively high safety rate and reward. In addition, we compare our method to the CAM without the graph decomposition. We observe that the graph decomposition works effectively in both environments. The performance of the Car environment benefits more from the graph decomposition. This advancement is because the density of Car agents in 2D is higher than that of the Drone agent in 3D. The higher density makes the test task more unlike the training task, which explains why the graph decomposition is more beneficial for Car.

%% file: main.bbl
\begin{thebibliography}{10}

\bibitem{manufact1}
Jos{\'e} Barbosa, Paulo Leit{\~a}o, Emmanuel Adam, and Damien Trentesaux.
\newblock Dynamic self-organization in holonic multi-agent manufacturing
  systems: The adacor evolution.
\newblock {\em Computers in industry}, 66:99--111, 2015.

\bibitem{manufact2}
Ray~Y Zhong, Xun Xu, Eberhard Klotz, and Stephen~T Newman.
\newblock Intelligent manufacturing in the context of industry 4.0: a review.
\newblock {\em Engineering}, 3(5):616--630, 2017.

\bibitem{transport1}
Ekim Yurtsever, Jacob Lambert, Alexander Carballo, and Kazuya Takeda.
\newblock A survey of autonomous driving: Common practices and emerging
  technologies.
\newblock {\em IEEE access}, 8:58443--58469, 2020.

\bibitem{survei1}
J{\"o}rg~P M{\"u}ller and Klaus Fischer.
\newblock Application impact of multi-agent systems and technologies: A survey.
\newblock In {\em Agent-oriented software engineering}, pages 27--53. Springer,
  2014.

\bibitem{masurvey1}
Reza Olfati-Saber, J~Alex Fax, and Richard~M Murray.
\newblock Consensus and cooperation in networked multi-agent systems.
\newblock {\em Proceedings of the IEEE}, 95(1):215--233, 2007.

\bibitem{masurvey2}
Yongcan Cao, Wenwu Yu, Wei Ren, and Guanrong Chen.
\newblock An overview of recent progress in the study of distributed
  multi-agent coordination.
\newblock {\em IEEE Transactions on Industrial informatics}, 9(1):427--438,
  2012.

\bibitem{panait2005scalability}
Liviu Panait and Sean Luke.
\newblock Cooperative multi-agent learning: The state of the art.
\newblock {\em Autonomous agents and multi-agent systems}, 11(3):387--434,
  2005.

\bibitem{brambilla2013scalability}
Manuele Brambilla, Eliseo Ferrante, Mauro Birattari, and Marco Dorigo.
\newblock Swarm robotics: a review from the swarm engineering perspective.
\newblock {\em Swarm Intelligence}, 7(1):1--41, 2013.

\bibitem{macommunicate}
Jakob Foerster, Ioannis~Alexandros Assael, Nando De~Freitas, and Shimon
  Whiteson.
\newblock Learning to communicate with deep multi-agent reinforcement learning.
\newblock {\em Advances in neural information processing systems}, 29, 2016.

\bibitem{gpg}
Arbaaz Khan, Ekaterina Tolstaya, Alejandro Ribeiro, and Vijay Kumar.
\newblock Graph policy gradients for large scale robot control.
\newblock In {\em Conference on robot learning}, pages 823--834. PMLR, 2020.

\bibitem{PIC}
Iou-Jen Liu, Raymond~A Yeh, and Alexander~G Schwing.
\newblock Pic: permutation invariant critic for multi-agent deep reinforcement
  learning.
\newblock In {\em Conference on Robot Learning}, pages 590--602. PMLR, 2020.

\bibitem{maddpg}
Ryan Lowe, Yi~Wu, Aviv Tamar, Jean Harb, Pieter Abbeel, and Igor Mordatch.
\newblock Multi-agent actor-critic for mixed cooperative-competitive
  environments.
\newblock In Isabelle Guyon, Ulrike von Luxburg, Samy Bengio, Hanna~M. Wallach,
  Rob Fergus, S.~V.~N. Vishwanathan, and Roman Garnett, editors, {\em Advances
  in Neural Information Processing Systems 30: Annual Conference on Neural
  Information Processing Systems 2017, December 4-9, 2017, Long Beach, CA,
  {USA}}, pages 6379--6390, 2017.

\bibitem{QMIX}
Tabish Rashid, Mikayel Samvelyan, Christian Schroeder, Gregory Farquhar, Jakob
  Foerster, and Shimon Whiteson.
\newblock Qmix: Monotonic value function factorisation for deep multi-agent
  reinforcement learning.
\newblock In {\em International Conference on Machine Learning}, pages
  4295--4304. PMLR, 2018.

\bibitem{rewardhacking1}
Dario Amodei, Chris Olah, Jacob Steinhardt, Paul Christiano, John Schulman, and
  Dan Man{\'e}.
\newblock Concrete problems in ai safety.
\newblock {\em arXiv preprint arXiv:1606.06565}, 2016.

\bibitem{rewardhacking2}
Dylan Hadfield{-}Menell, Smitha Milli, Pieter Abbeel, Stuart~J. Russell, and
  Anca~D. Dragan.
\newblock Inverse reward design.
\newblock In Isabelle Guyon, Ulrike von Luxburg, Samy Bengio, Hanna~M. Wallach,
  Rob Fergus, S.~V.~N. Vishwanathan, and Roman Garnett, editors, {\em Advances
  in Neural Information Processing Systems 30: Annual Conference on Neural
  Information Processing Systems 2017, December 4-9, 2017, Long Beach, CA,
  {USA}}, pages 6765--6774, 2017.

\bibitem{stern2019multi}
Roni Stern, Nathan~R Sturtevant, Ariel Felner, Sven Koenig, Hang Ma, Thayne~T
  Walker, Jiaoyang Li, Dor Atzmon, Liron Cohen, TK~Satish Kumar, et~al.
\newblock Multi-agent pathfinding: Definitions, variants, and benchmarks.
\newblock In {\em Twelfth Annual Symposium on Combinatorial Search}, 2019.

\bibitem{pbs}
Hang Ma, Daniel Harabor, Peter~J Stuckey, Jiaoyang Li, and Sven Koenig.
\newblock Searching with consistent prioritization for multi-agent path
  finding.
\newblock In {\em Proceedings of the AAAI Conference on Artificial
  Intelligence}, volume~33, pages 7643--7650, 2019.

\bibitem{cbs}
Guni Sharon, Roni Stern, Ariel Felner, and Nathan~R Sturtevant.
\newblock Conflict-based search for optimal multi-agent pathfinding.
\newblock {\em Artificial Intelligence}, 219:40--66, 2015.

\bibitem{qingbiao1}
Qingbiao Li, Fernando Gama, Alejandro Ribeiro, and Amanda Prorok.
\newblock Graph neural networks for decentralized multi-robot path planning.
\newblock In {\em 2020 IEEE/RSJ International Conference on Intelligent Robots
  and Systems (IROS)}, pages 11785--11792, 2020.

\bibitem{qingbiao2}
Qingbiao Li, Weizhe Lin, Zhe Liu, and Amanda Prorok.
\newblock Message-aware graph attention networks for large-scale multi-robot
  path planning.
\newblock {\em IEEE Robotics and Automation Letters}, 6(3):5533--5540, 2021.

\bibitem{primal}
Guillaume Sartoretti, Justin Kerr, Yunfei Shi, Glenn Wagner, TK~Satish Kumar,
  Sven Koenig, and Howie Choset.
\newblock Primal: Pathfinding via reinforcement and imitation multi-agent
  learning.
\newblock {\em IEEE Robotics and Automation Letters}, 4(3):2378--2385, 2019.

\bibitem{orca}
Jur van~den Berg, Stephen~J Guy, Ming Lin, and Dinesh Manocha.
\newblock Reciprocal n-body collision avoidance.
\newblock In {\em Robotics research}, pages 3--19. Springer, 2011.

\bibitem{bandyopadhyay2017probabilistic}
Saptarshi Bandyopadhyay, Soon-Jo Chung, and Fred~Y Hadaegh.
\newblock Probabilistic and distributed control of a large-scale swarm of
  autonomous agents.
\newblock {\em IEEE Transactions on Robotics}, 33(5):1103--1123, 2017.

\bibitem{potential86}
Oussama Khatib.
\newblock Real-time obstacle avoidance for manipulators and mobile robots.
\newblock In {\em Autonomous robot vehicles}, pages 396--404. Springer, 1986.

\bibitem{Barraquand1992APF}
Jerome Barraquand, Bruno Langlois, and J-C Latombe.
\newblock {Numerical potential field techniques for robot path planning}.
\newblock {\em IEEE transactions on systems, man, and cybernetics},
  22(2):224--241, 1992.

\bibitem{Ge2002APF}
Shuzhi~Sam Ge and Yun~J Cui.
\newblock {Dynamic motion planning for mobile robots using potential field
  method}.
\newblock {\em Autonomous robots}, 13(3):207--222, 2002.

\bibitem{hernandez2011apf}
Eduardo~G Hern{\'a}ndez-Mart{\'\i}nez, Eduardo Aranda-Bricaire, and
  F~Alkhateeb.
\newblock {\em Convergence and collision avoidance in formation control: A
  survey of the artificial potential functions approach}.
\newblock INTECH Open Access Publisher Rijeka, Croatia, 2011.

\bibitem{glas}
Benjamin Riviere, Wolfgang H{\"o}nig, Yisong Yue, and Soon-Jo Chung.
\newblock Glas: Global-to-local safe autonomy synthesis for multi-robot motion
  planning with end-to-end learning.
\newblock {\em IEEE Robotics and Automation Letters}, 5(3):4249--4256, 2020.

\bibitem{neuralswarm}
Guanya Shi, Wolfgang H{\"o}nig, Yisong Yue, and Soon-Jo Chung.
\newblock Neural-swarm: Decentralized close-proximity multirotor control using
  learned interactions.
\newblock In {\em 2020 IEEE International Conference on Robotics and Automation
  (ICRA)}, pages 3241--3247. IEEE, 2020.

\bibitem{macbf}
Zengyi Qin, Kaiqing Zhang, Yuxiao Chen, Jingkai Chen, and Chuchu Fan.
\newblock {Learning Safe Multi-Agent Control with Decentralized Neural Barrier
  Certificates}.
\newblock In {\em Conference on Learning Representations}. Conference on
  Learning Representations, jan 2021.

\bibitem{ames2019control}
Aaron~D Ames, Samuel Coogan, Magnus Egerstedt, Gennaro Notomista, Koushil
  Sreenath, and Paulo Tabuada.
\newblock Control barrier functions: Theory and applications.
\newblock In {\em 2019 18th European Control Conference (ECC)}, pages
  3420--3431, Naples, Italy, June 2019.

\bibitem{DBLP:conf/amcc/HanHP18}
Dongkun Han, Lixing Huang, and Dimitra Panagou.
\newblock Approximating the region of multi-task coordination via the optimal
  lyapunov-like barrier function.
\newblock In {\em 2018 Annual American Control Conference, {ACC} 2018,
  Milwaukee, WI, USA, June 27-29, 2018}, pages 5070--5075. {IEEE}, 2018.

\bibitem{wan2017semi}
Shuo Wan, Jiaxun Lu, and Pingyi Fan.
\newblock Semi-centralized control for multi robot formation.
\newblock In {\em 2017 2nd International Conference on Robotics and Automation
  Engineering (ICRAE)}, pages 31--36. IEEE, 2017.

\bibitem{GCN}
Thomas~N. Kipf and Max Welling.
\newblock Semi-supervised classification with graph convolutional networks.
\newblock In {\em 5th International Conference on Learning Representations,
  {ICLR} 2017, Toulon, France, April 24-26, 2017, Conference Track
  Proceedings}. OpenReview.net, 2017.

\bibitem{MPNN}
Justin Gilmer, Samuel~S Schoenholz, Patrick~F Riley, Oriol Vinyals, and
  George~E Dahl.
\newblock Neural message passing for quantum chemistry.
\newblock In {\em International conference on machine learning}, pages
  1263--1272. PMLR, 2017.

\bibitem{interactionn}
Peter~W. Battaglia, Razvan Pascanu, Matthew Lai, Danilo~Jimenez Rezende, and
  Koray Kavukcuoglu.
\newblock Interaction networks for learning about objects, relations and
  physics.
\newblock In Daniel~D. Lee, Masashi Sugiyama, Ulrike von Luxburg, Isabelle
  Guyon, and Roman Garnett, editors, {\em Advances in Neural Information
  Processing Systems 29: Annual Conference on Neural Information Processing
  Systems 2016, December 5-10, 2016, Barcelona, Spain}, pages 4502--4510, 2016.

\bibitem{gnnancestor}
Franco Scarselli, Marco Gori, Ah~Chung Tsoi, Markus Hagenbuchner, and Gabriele
  Monfardini.
\newblock The graph neural network model.
\newblock {\em IEEE transactions on neural networks}, 20(1):61--80, 2008.

\bibitem{deepset}
Manzil Zaheer, Satwik Kottur, Siamak Ravanbakhsh, Barnabas Poczos, Russ~R
  Salakhutdinov, and Alexander~J Smola.
\newblock Deep sets.
\newblock {\em Advances in neural information processing systems}, 30, 2017.

\bibitem{pointnet}
Charles~Ruizhongtai Qi, Hao Su, Kaichun Mo, and Leonidas~J. Guibas.
\newblock Pointnet: Deep learning on point sets for 3d classification and
  segmentation.
\newblock In {\em 2017 {IEEE} Conference on Computer Vision and Pattern
  Recognition, {CVPR} 2017, Honolulu, HI, USA, July 21-26, 2017}, pages 77--85.
  {IEEE} Computer Society, 2017.

\bibitem{gnnmp}
Arbaaz Khan, Alejandro Ribeiro, Vijay Kumar, and Anthony~G. Francis.
\newblock Graph neural networks for motion planning.
\newblock {\em CoRR}, abs/2006.06248, 2020.

\bibitem{chenning2021collision}
Chenning Yu and Sicun Gao.
\newblock Reducing collision checking for sampling-based motion planning using
  graph neural networks.
\newblock In {\em Proceedings of the 35rd International Conference on Neural
  Information Processing Systems}, 2021.

\bibitem{prorok1}
Lifeng Zhou, Vishnu~D Sharma, Qingbiao Li, Amanda Prorok, Alejandro Ribeiro,
  and Vijay Kumar.
\newblock Graph neural networks for decentralized multi-robot submodular action
  selection.
\newblock {\em arXiv preprint arXiv:2105.08601}, 2021.

\bibitem{tolstaya2020multi}
Ekaterina Tolstaya, James Paulos, Vijay Kumar, and Alejandro Ribeiro.
\newblock Multi-robot coverage and exploration using spatial graph neural
  networks.
\newblock In {\em 2021 IEEE/RSJ International Conference on Intelligent Robots
  and Systems (IROS)}, pages 8944--8950. IEEE, 2020.

\bibitem{superglue}
Paul-Edouard Sarlin, Daniel DeTone, Tomasz Malisiewicz, and Andrew Rabinovich.
\newblock {SuperGlue}: Learning feature matching with graph neural networks.
\newblock In {\em CVPR}, 2020.

\bibitem{alexnet}
Alex Krizhevsky, Ilya Sutskever, and Geoffrey~E. Hinton.
\newblock Imagenet classification with deep convolutional neural networks.
\newblock In Peter~L. Bartlett, Fernando C.~N. Pereira, Christopher J.~C.
  Burges, L{\'{e}}on Bottou, and Kilian~Q. Weinberger, editors, {\em Advances
  in Neural Information Processing Systems 25: 26th Annual Conference on Neural
  Information Processing Systems 2012. Proceedings of a meeting held December
  3-6, 2012, Lake Tahoe, Nevada, United States}, pages 1106--1114, 2012.

\bibitem{resnet}
Kaiming He, Xiangyu Zhang, Shaoqing Ren, and Jian Sun.
\newblock Deep residual learning for image recognition.
\newblock {\em CoRR}, abs/1512.03385, 2015.

\bibitem{resgnn}
Binxuan Huang and Kathleen~M Carley.
\newblock Residual or gate? towards deeper graph neural networks for inductive
  graph representation learning.
\newblock {\em arXiv preprint arXiv:1904.08035}, 2019.

\bibitem{sutton2018reinforcement}
Richard~S Sutton and Andrew~G Barto.
\newblock {\em Reinforcement learning: An introduction}.
\newblock MIT press, 2018.

\bibitem{rudin1976principles}
Walter Rudin et~al.
\newblock {\em Principles of mathematical analysis}, volume~3.
\newblock McGraw-hill New York, 1976.

\bibitem{pybullet}
Erwin Coumans and Yunfei Bai.
\newblock Pybullet, a python module for physics simulation for games, robotics
  and machine learning.
\newblock 2016.

\bibitem{dubinscar}
Lester~E Dubins.
\newblock On curves of minimal length with a constraint on average curvature,
  and with prescribed initial and terminal positions and tangents.
\newblock {\em American Journal of mathematics}, 79(3):497--516, 1957.

\bibitem{dawson2022safe}
Charles Dawson, Zengyi Qin, Sicun Gao, and Chuchu Fan.
\newblock Safe nonlinear control using robust neural lyapunov-barrier
  functions.
\newblock In {\em Conference on Robot Learning}, pages 1724--1735. PMLR, 2022.

\bibitem{ddpg}
Timothy~P Lillicrap, Jonathan~J Hunt, Alexander Pritzel, Nicolas Heess, Tom
  Erez, Yuval Tassa, David Silver, and Daan Wierstra.
\newblock Continuous control with deep reinforcement learning.
\newblock {\em arXiv preprint arXiv:1509.02971}, 2015.

\end{thebibliography}
